\documentclass{article} 
\usepackage{iclr2026_conference,times}

\usepackage{amsmath,amsfonts,bm}

\def\eqref#1{equation~\ref{#1}}

\def\1{\bm{1}}

\DeclareMathAlphabet{\mathsfit}{\encodingdefault}{\sfdefault}{m}{sl}
\SetMathAlphabet{\mathsfit}{bold}{\encodingdefault}{\sfdefault}{bx}{n}

\usepackage{hyperref}
\usepackage{url}
\usepackage{graphicx}

\title{DeCAL Tokenwise Compression} 

\author{Sameer Panwar  \\
Google DeepMind \\
\texttt{sameerpanwar@google.com} \\
}

\iclrfinalcopy 
\begin{document}

\maketitle

\begin{abstract}
This paper introduces DeCAL, a new method for tokenwise compression.  DeCAL uses an encoder-decoder language model pretrained with denoising to learn to produce high-quality, general-purpose compressed representations from the encoder.  DeCAL applies small modifications to the encoder, with the emphasis on maximizing compression quality, even at the expense of compute.  We show that DeCAL at 2x compression can match uncompressed on several downstream tasks, with usually only a minor dropoff in metrics up to 8x compression, among question-answering, summarization, and multi-vector retrieval tasks.  DeCAL offers significant savings where pre-computed dense representations can be utilized, and we believe the approach can be further developed to be more broadly applicable.
\end{abstract}

\section{Introduction}

Transformers \citep{Vaswani_NIPS2017_3f5ee243} have proven very effective at a broad range of natural language tasks, but inference costs increase dramatically as sequence lengths increase.  Efforts to address the cost often focus on speeding up the attention block \citep{transformers_survey}, the main bottleneck, usually by sparsifying compute, but retaining all tokens since it is unknown which subsequent attention computations might attend to them.  Pruning input tokens \citep{pruning_survey} is another strategy, but that leads to forgetting, which may sacrifice performance on certain tasks.

Alternatively, we can compress the input sequence by merging multiple token representations together, potentially retaining more information.  Intuitively, token sequences should be fairly compressible given they are built from subwords and other predictable patterns.   We know tokens contain nonuniform amounts of information; punctuation, for example, has less content than a long sentencepiece \citep{sentencepiece}.  Several subwords individually do not contain much information, but in sequence together they do; however, that information does not require their original space.  Hence, the opportunity for compression is clear.

Approaches for tokenwise compression rely on variants of attention pooling (as in Funnel Transformer \citep{funnel}) and convolution (as in CANINE \citep{canine}), primarily to deliver substantial training and inference speedups.  The speedups do come at a cost to model performance, as we have observed in practice with the Funnel Transformer.  

More recent work such as AutoCompressor \citep{chevalier2023adapting}, COCOM \citep{cocom}, and ICAE \citep{ge2024incontext} share the goal of LLM context compression.  They all derive compressive encoders from decoder-only LLMs in order to fit more information within the context budget of the LLM, as well as save inference time when the context chunks can be precomputed.

This work focuses on maximizing compressed representation quality, without making tradeoffs for immediate inference speedup, leaving compute optimization for later.  We utilize the T5 framework, which natively supports encoder-decoder models, in order to readily explore the design space including compressive pretraining from scratch.  This, however, makes it difficult to make apples-to-apples comparisons against decoder-only-derived LLM approaches.  Instead, we build a proxy from the common elements of the recent LLM context compressors and compare against it, trained and evaluated identically in our environment.

We present DeCAL, a novel method for tokenwise compression.  DeCAL employs an encoder structure in which we prepend an input-derived latent sequence to the encoder input.  This shorter latent sequence successively attends to the evolving input sequence, layer by layer, but only the latent is output.  This structure forms the basis of the name DeCAL, short for Deep Cross-Attended Latents.

Our main contributions are as follows:
\begin{itemize}
\item  We train DeCAL (as an encoder-decoder language model) from scratch with compression on a denoising language modeling task.  We show that this is superior to the more typical approach of autoencoding (mixed with text continuation) after standard pretraining.
\item We initialize DeCAL's latent sequence with pooled input tokens combined with a learnable embedding vector, and show that this outperforms using the learnable vector alone.
\item A simple alternative to DeCAL's prepending \(m\) latent tokens to the input sequence is to instead perform attention pooling on the output layer down to \(m\) tokens.  Though the latter is faster, we establish that it delivers inferior compression. 
\item  We fine-tune DeCAL on context-intensive tasks, namely summarization and question answering, and compare different levels of compression, with only minor metrics dropoff at 8x compression.  We also construct a proxy for typical LLM context compression approaches to compare against and determine that the proxy falls short of DeCAL performance. 
\item We illustrate the versatility of DeCAL by additionally applying it to multi-vector retrieval tasks.
\end{itemize}

\section{Related Work}

There are a number of studies that involve compressing token sequences, though most do this internally for efficiency, not as an explicit usable output sequence.  

\textbf{Compressive Encoder-only.}  The Funnel Transformer \citep{funnel} applies multiple steps of attention-pooling compression, followed by a single step of upsampling back to full sequence length (plus residual from the last full length layer).  The upsampled layers are used for masked language pretraining, then discarded.  In contrast, DeCAL does not compress the input stepwise in-line to the output, but rather cumulatively, alongside the input (as detailed under Methods).  DeCAL uses a full-size decoupled autoregressive decoder to interpret the encoder’s compressed output and so more naturally handles more complex language modeling such as variable-length corrupted token spans.

Perceiver \citep{perceiver} uses a short latent sequence that repeatedly cross-attends to the original input sequence, thus requiring much less compute.  DeCAL instead combines the latent with the original input and both sequences evolve layer by layer, requiring net additional compute (trading off for higher compression quality).

\textbf{Internal compression.}  Several studies use compression internally, but not in any externally usable form.  Hourglass \citep{hourglass} is a decoder-only architecture that resembles Funnel Transformer, with successive compression steps, but these are later followed by successive upsampling steps each with residual connections.  This is done for the purpose of efficiency for long sequences.

CANINE \citep{canine} applies convolutions for both downsampling and upsampling (but no residual connections), with the goal of handling character-level inputs.  Byte Latent Transformer \citep{pagnoni2024bytelatenttransformerpatches} groups input bytes into variable length patches which undergo attention pooling into one global token per patch before being upsampled with reversed attention pooling (the last full-length layer output serves as the attention queries).  Both approaches apply the bulk of their compute in their compressed inner layers.

LongT5’s \citep{guo-etal-2022-longt5} sparse attention compresses each layer’s input using average pooling to serve as ephemeral KV sources for long-range attention.

\textbf{History compression.}  Additional studies compress past history of decoders to cope with very long contexts.  The Compressive Transformer \citep{compressive_transformer} propagates past token history through recurrent encoding of token blocks, and then extends its memory capacity by compressing (via convolution) all but the last block.  The Recurrent Memory Transformer \citep{recurrent_memory} also performs recurrent encoding of token blocks, but carries over only a small fixed number of tokens per block, for very high compression to achieve 1M+ token contexts, but is very task-specific.

\textbf{Pruning.}  A less common technique to shorten sequences is pruning tokens, such as PoWER-BERT \citep{power_bert}, which drops a few tokens every layer, knowing that at the output, only the CLS token is used.  However, this is only added after pretrain and fine-tune, then an auxiliary model is used to identify how to drop tokens, resulting in a task-specific inference speedup.  NUGGET \citep{nugget} has an encoder-decoder format, but adds small FFNs on the encoder outputs to perform top-K selection.  NUGGET is trained as an autoencoder, but is limited to 128 token blocks.

\textbf{Soft prompts.}  Soft prompts \citep{prefix_tuning} involve adding a learnable prefix to a frozen model’s input, using a different prefix per task as an alternative to separate fine-tuning per task.  This looks superficially like DeCAL in its learnable prefix, but is functionally orthogonal; DeCAL’s encoder outputs only the latent sequence, which is a function of the input sequence, unlike soft prompts.

\textbf{Context compression.}  AutoCompressors \citep{chevalier2023adapting} build on the Recurrent Memory Transformer \citep{recurrent_memory} by keeping prior blocks' summary vectors and successively concatenating them to finally obtain an effective compressed form that is used as a soft prompt.  The In-context Autoencoder (ICAE) \citep{ge2024incontext} works similarly, but compresses blocks in parallel instead of recursively, so blocks do not see their neighboring context.  ICAE's compression training consists of autoencoding with text continuation, in contrast to Autocompressor's next token prediction.  COCOM \citep{cocom} develops further by unfreezing the decoder, and achieves a better compression ratio than ICAE, but encodes smaller chunks.

All of these works share the common approach of appending special context tokens to accumulate the compressed output representation, which is similar to DeCAL, though DeCAL initializes the latent tokens differently. All adapt from decoder-only LLMs, and do not apply denoising during their compression training, unlike DeCAL.

\medskip
It is worth observing the commonality that the underlying compression operation in all of the above techniques is most often cross-attention from a small sequence to a larger one and, in some cases, convolution.  

\section{Methods}

The essential elements of DeCAL are:
\begin{itemize}
\item An encoder-decoder model configuration
\item A compressive encoder structure with latent sequence prepended to the input
\item Pooling of input tokens to seed the initial latent sequence state
\item A denoising language modeling pretrain task
\end{itemize}

In an encoder-decoder language model, as in T5 \citep{t5}, the encoder provides the input context, while the decoder makes next-token predictions.  This provides two very useful properties: 1) Unlike encoder-only masked-LM \citep{devlin-etal-2019-bert}, there are no indicators of how many tokens a compressed input token originated from, nor any residuals of the uncompressed input; the decoder is cleanly decoupled and must discern the contents of each compressed token.  This puts more weight on the encoder to do a better job of compression and on the decoder to validate that.  2) Training examples have separate sequences for the encoder and decoder, allowing us to tune the amount of work each must do.  

We find that the standard span corruption language modeling task in T5 \citep{t5} works very well.  For the encoder input, it removes spans of tokens, replacing them with sentinel tokens.  The decoder target sequence is the complement, with just the tokens that had been removed, and sentinel tokens between them representing the token spans provided by the encoder.  So despite the input to the decoder being compressed, it must still identify the sentinel tokens and the preceding and succeeding tokens, then denoise to output the correct span.

\begin{figure}[h]
    \centering
    \includegraphics[width=0.5\linewidth]{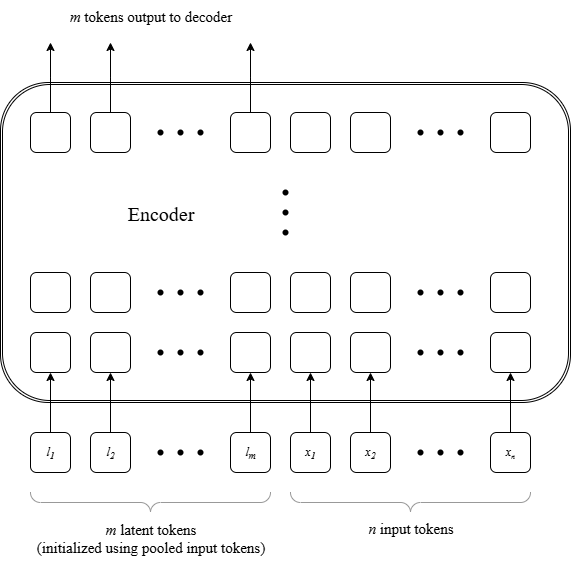}
    \caption{Diagram of DeCAL encoder input and output for compressing \textit{n} input tokens to \textit{m} output tokens.}
    \label{fig:1}
\end{figure}

DeCAL is constructed as follows, using a standard T5 encoder-decoder:
\begin{itemize}
\item We take $\textbf{\textit{x}} = [x_1, ..., x_n]$ as the input embeddings to be compressed.
\item To create the initial latent sequence $\textbf{\textit{l}} = [l_1, ..., l_m]$ where \(m < n\) (giving us a compression ratio \(C = n/m\)),
\begin{itemize}
 \item We start with a learnable vector \textbf{\textit{v}} repeated \textit{m} times. 
 \item We then add the input embeddings, pooled from length \textit{n} to \textit{m} by pooling function \textit{P}.  In this work, for \textit{P}, we use a strided mean pool with window size \textit{C} (same as Funnel Transformer’s \citep{funnel} attention query in its attention pooling).
 \item Lastly, we apply a layer norm.
 \item Thus \(\textbf{\textit{l}} = LayerNorm(\textbf{\textit{v}}_{\times m} + P(\textbf{\textit{x}}))\)
 \end{itemize}
\item The input to the encoder is \(\textbf{\textit{l}} \oplus \textbf{\textit{x}}\), thus longer than the original input, as shown in Figure~\ref{fig:1}.
\item The encoder is internally unmodified, so there are negligible additional parameters.  Note that both the \textbf{\textit{l}} and \textit{\textbf{x}} subsequences are able to attend to each other as their representations are updated in each layer.
\item We assign the position index of tokens in \textbf{\textit{l}} to be the integer mean of the position indices of the corresponding tokens in \textbf{\textit{x}}.  This is used for T5 relative position calculations.
\item The encoder output is exclusively the \textit{m} output embeddings corresponding to \textbf{\textit{l}}, as shown in Figure~\ref{fig:1}.  
\item The decoder is unchanged.
\end{itemize}

In this scheme, we are logically assigning each token in \textbf{\textit{l}} to a span of tokens in \textbf{\textit{x}}, forming the correspondence of which \textbf{\textit{x}} tokens we expect to be compressed into which latent token.  We initialize \textbf{\textit{l}} on this basis, but note we do nothing further with this correspondence, such as attention masking, etc.  In our experiments, we only use a constant C, but the design can easily accommodate variable length mappings, such as used in BLT \citep{pagnoni2024bytelatenttransformerpatches}.  \textbf{\textit{l}} starts only from initial token embeddings, but as the \textbf{\textit{x}} subsequence is updated after each transformer layer, the \textbf{\textit{l}} subsequence can attend to \textbf{\textit{x}} as well as itself, building up a compressed contextualized representation.  Likewise \textbf{\textit{x}} is able to attend to \textbf{\textit{l}}, though we haven’t tested its utility.  We observe that this approach increases compute cost due to the increase in encoder input length; interestingly, this means compressing more (shorter \textbf{\textit{l}}) is lower compute than compressing less (longer \textbf{\textit{l}}), which has to be taken into account when trading off compute against compression quality.

The encoder maintains constant compression C, scaling \textit{m} as necessary if \textit{n} changes; e.g. if after pretraining at 4x and 1024 token inputs for example, we then fine-tune the model on a task with 8192 token inputs, the encoder will output 2048 tokens.

It would seem logical that performing compression alongside a normal transformer stack would require differing parameters; we tried having separate parameters for the transformer blocks handling the latent tokens, however, this did not produce better results, so further tweaking may be required to continue down that path.

\section{Experiments}

\subsection{Setup}

All our experiments used the T5.1.1\footnote{\url{https://github.com/google-research/text-to-text-transfer-transformer/blob/main/released_checkpoints.md\#t511}} Large encoder-decoder, pretrained from scratch on 128M examples from C4 with a sequence length of 1024, and the default span corruption configuration of 15\% corruption, which means 85\% of the training tokens are fed to the encoder and 15\% to the decoder.  We used 10k steps of linear warmup and used a 4x larger relative position \texttt{num\_buckets} and \texttt{max\_distance} than T5’s default, since that worked best.  The DeCAL models were trained identically, with the model structure as described under Methods and compression ratios varying from 2x to 8x.

\subsection{Encoder-Decoder Experiments}

\subsubsection{Datasets}
Since this work is focused on compressing encoder inputs, we focused on summarization and question-answering tasks, which are naturally context-intensive, to evaluate compression effectiveness.  For summarization, we used the arXiv \citep{arxiv_pubmed_sum} and PubMed \citep{arxiv_pubmed_sum} datasets, for which the document is the encoder input and the abstract is the target summary.  We also included CNN / DailyMail \citep{cnnsum}, where the news article is the encoder input and the article's summary bullets are the target summary, as well as MediaSum \citep{zhu-etal-2021-mediasum}, where interview transcripts are the encoder input and their topic and overviews served as the target summary.

For question-answering, we used HotpotQA \citep{yang-etal-2018-hotpotqa} and TriviaQA \citep{joshi-etal-2017-triviaqa}, for which the question and article context are given to the encoder, and the (first) answer is the target.  We fine-tuned on 6M examples, except arXiv, for which we trained on 12M examples, since that task took longer to converge.  We limited context to 4096 tokens for all tasks.

\subsubsection{Language Modeling}
We tested DeCAL with compression ratios of 2x, 4x, and 8x.  Since these ratios get quite aggressive, it can be instructive to see how these impact encoder-decoder pretraining.  As shown in Table~\ref{table:1}, the DeCAL target token prediction accuracy at 2x is able to match the (uncompressed) T5 Large baseline, while 4x is sufficiently close that further training may close much of the gap.  At 8x, accuracy takes another comparable step down.  Training at very high compression ratios of 16x and higher proved unstable during pretraining.

\begin{table}[h]
\renewcommand{\arraystretch}{1.3}
\begin{center}
\begin{tabular}{c | c} 
 \hline
 & Accuracy \\ 
 \hline \
 Baseline & 70.2 \\ 
 \hline
 DeCAL 2x & \textbf{70.4} \\ 
 DeCAL 4x & 69.2 \\ 
 DeCAL 8x & 68.0 \\ 
 DeCAL 16x & unstable \\
 \hline
\end{tabular}
\caption{Span corruption pretraining accuracy for different levels of DeCAL compression against no compression.}
\label{table:1}
\end{center}
\end{table}

\subsubsection{Summarization}
Table~\ref{table:2} shows ROUGE metrics for the summarization datasets.  As with pretraining, we see that DeCAL 2x is always comparable to the baseline, while 4x is only modestly behind.  8x is a notable step lower, but not significantly considering the decoder only cross-attends to 512 tokens versus 4096 uncompressed.  We speculate DeCAL compression does favorably here because enough of the general semantics of the full input is preserved for the purpose of abstractive summarization.

\begin{table}[h]
\renewcommand{\arraystretch}{1.3}
\begin{center}
\begin{tabular}{c | c | c | c | c | c | c} 
 \hline
 \multicolumn{1}{c|}{} & \multicolumn{3}{c|}{arXiv} & \multicolumn{3}{c}{PubMed} \\
 \cline{2-7}
  & R-1 & R-2 & R-L & R-1 & R-2 & R-L \\
 \hline 
 Baseline & \textbf{45.6} & \textbf{18.9} & \textbf{41.6} & 48.5 & \textbf{23.1} & 45.1 \\
 \hline
 DeCAL 2x & 45.3 & \textbf{18.9} & 41.4 & \textbf{48.6} & \textbf{23.1} & \textbf{45.2} \\
 DeCAL 4x & 44.9 & 18.6 & 41.0 & 47.9 & 22.5 & 44.5 \\
 DeCAL 8x & 44.0 & 17.9 & 40.1 & 47.2 & 21.4 & 43.8 \\
 \hline

 \hline
 \multicolumn{1}{c|}{} & \multicolumn{3}{c|}{CNN / DailyMail} & \multicolumn{3}{c}{MediaSum} \\
 \cline{2-7}
  & R-1 & R-2 & R-L & R-1 & R-2 & R-L \\
 \hline 
 Baseline & 42.9 & 20.7 & 37.3 & \textbf{35.1} & \textbf{18.5} & \textbf{32.1} \\
 \hline
 DeCAL 2x & \textbf{43.2} & \textbf{20.9} & \textbf{37.5} & 35.0 & 18.3 & 32.0\\
 DeCAL 4x & 42.2 & 20.1 & 36.7 & 34.1 & 17.6 & 31.2 \\
 DeCAL 8x & 41.4 & 19.1 & 35.9 & 32.2 & 16.2 & 29.8 \\
 \hline

\end{tabular}%
\caption{ROUGE metrics on summarization tasks for different levels of DeCAL compression against no compression.}
\label{table:2}
\end{center}
\end{table}

\subsubsection{Question-Answering}
For the question-answering datasets, presented in Table~\ref{table:3}, we see again that DeCAL 2x can fully match the baseline.  DeCAL 4x and 8x gradually drop off, but much more so for HotpotQA, likely due to its multi-hop reasoning characteristic.  This range of compression ratios gives us a way to trade off compression versus model performance, all the way to 8x.

\begin{table}[h]
\renewcommand{\arraystretch}{1.3}
\begin{center}
\begin{tabular}{c | c | c | c | c} 
 \multicolumn{1}{c|}{} & \multicolumn{2}{c|}{TriviaQA} & \multicolumn{2}{c}{HotpotQA} \\
 \cline{2-5}
  & EM & F1 & EM & F1 \\
 \hline \
 Baseline & 76.8 & 79.1 & 65.1 & 79.2 \\
 \hline
 DeCAL 2x & \textbf{77.0} & \textbf{79.2} & \textbf{65.2} & \textbf{79.5} \\
 DeCAL 4x & 76.0 & 78.3 & 62.9 & 77.0 \\
 DeCAL 8x & 75.1 & 77.2 & 60.3 & 74.4 \\
 \hline
\end{tabular}
\caption{SQuAD metrics on question-answering tasks for different levels of DeCAL compression against no compression.}
\label{table:3}
\end{center}
\end{table}

\subsubsection{Comparisons}
Having a basis of comparison with prior work is useful in placing new work in context.  The works that are most similar to DeCAL are LLM context compressors.  However, these are built on a myriad of frameworks, have varying model sizes, maximum chunk lengths, and evaluation tasks.  We feel it is more beneficial to instead extract the common elements of those designs to construct a proxy of these current approaches and compare against it.  This eliminates extraneous effects from arbitrary differences, such as training data, token limits, LLM input packing strategies, etc.

Using AutoCompressor \citep{chevalier2023adapting}, COCOM \citep{cocom}, and ICAE \citep{ge2024incontext} as reference, the common key elements are as follows:
\begin{itemize}
    \item These all use learnable special tokens appended to the input to aggregate the compressed information.
    \item COCOM and ICAE both use autoencoding mixed 1:1 with text continuation for compression training.  We use C4 for these, for 26M examples, and start from our baseline (uncompressed) model.
    \item These all derive from a decoder-only model, and so are subject to causal attention masks.
\end{itemize}
We will call the proxy built as above \textit{CCProxy}.

Separately, an obvious alternative to prepending a latent sequence to the input is attention pooling the encoder's final output. This has lower compute, so it is also worth comparing.  We refer to this alternative as \textit{AttnPool} in our results, and only test 2x compression.  We train it identically to the DeCAL experiments, and we use mean pool (stride 2) to initialize the query of the attention pool layer. 

\begin{table}[h]
\renewcommand{\arraystretch}{1.3}
\begin{center}
\begin{tabular}{c | c | c | c | c | c} 
 \hline
 \multicolumn{1}{c|}{} & \multicolumn{3}{c|}{MediaSum} & \multicolumn{2}{c}{HotpotQA} \\
 \cline{2-6}
  & R-1 & R-2 & R-L & EM & F1 \\
 \hline 
 Baseline & \textbf{35.1} & \textbf{18.5} & \textbf{32.1} & 65.1 & 79.2 \\
 \hline
 DeCAL 2x & 35.0 & 18.3 & 32.0 & \textbf{65.2} & \textbf{79.5} \\
 DeCAL 4x & 34.1 & 17.6 & 31.2 & 62.9 & 77.0 \\
 DeCAL 8x & 32.2 & 16.2 & 29.8 & 60.3 & 74.4 \\
 \hline
 CCProxy 2x & 33.6 & 17.2 & 30.5 & 59.2 & 73.6 \\ 
 CCProxy 4x & 33.1 & 16.6 & 29.9 & 56.9 & 71.0 \\
 CCProxy 8x & 31.9 & 15.5 & 29.0 & 55.1 & 69.0 \\
 \hline
 AttnPool 2x & 34.1 & 17.5 & 31.1 & 60.2 & 74.4 \\
 \hline

\end{tabular}%
\caption{ROUGE metrics on MediaSum and SQuAD metrics on HotpotQA for different levels of compression for both DeCAL and CCProxy against no compression and AttnPool 2x.}
\label{table:4}
\end{center}
\end{table}

Due to resource limitations, we chose one summarization and one question-answering task for our comparisons.  We selected MediaSum and HotpotQA, since they showed greatest sensitivity to compression ratio.

From table~\ref{table:4}, we see that CCProxy is consistently outperformed by DeCAL at the same compression ratio; in fact, for MediaSum, CCProxy largely performs similarly to DeCAL at twice its compression ratio.  For HotpotQA, this task proves more difficult with CCProxy 2x coming in worse than DeCAL 8x.  The spread in metrics between 2x and 8x is actually slightly smaller for CCProxy than for DeCAL.  We can attribute the difference in performance directly to the CCProxy elements outlined above.

AttnPool 2x comes very close to DeCAL 4x on MediaSum and close to DeCAL 8x on HotpotQA.  This is a strong indicator that the extra compute cost from adding the latent sequence in DeCAL does translate into superior compressed representations.

\subsection{Encoder-Only, Retrieval Experiments}

ColBERT's \citep{colbert} late interaction approach offers superior retrieval to single-vector by virtue of allowing each query token to interact with each passage token, but suffers from dramatically greater compute cost (a dot product for each interaction).  Pruning techniques have been proposed \citep{plaid,aligner} to address the compute cost, but pruning loses information and does not save storage cost.  In contrast, DeCAL does not drop any tokens, so information should be lost more gracefully as compression is increased, and by reducing the representation up-front, storage costs are reduced.  We do not have a setup for reproducing either pruning technique for comparison (left for future work), but we include these results to illustrate the versatility of DeCAL and compare against the uncompressed baseline.

For these experiments, we took an already pretrained (as described under Setup) DeCAL model at some compression C, discarded the decoder and fine-tuned the encoder in a dual encoder configuration using T5X Retrieval\footnote{\url{https://github.com/google-research/t5x_retrieval}}.  We utilized Chamfer similarity loss as in ColBERT \citep{colbert}.

\subsubsection{Datasets}
We used MS MARCO \citep{msmarco} as the retrieval fine-tuning dataset (for 12M examples).  For evaluation, we used a subset of the BEIR \citep{thakur2021beir} retrieval benchmark: FiQA-2018, TREC-COVID, Touché-2020, NFCorpus, and SciFact, and we report NDCG@10.  These (smaller) datasets were selected simply due to resource constraints we had for these experiments.

\subsubsection{Results}

\begin{table}[h]
\renewcommand{\arraystretch}{1.3}
\begin{center}
\begin{tabular}{c | c | c | c | c | c} 
 \hline
 & FiQA & SciFact & NFCorpus & Touché & COVID \\
 \cline{2-6}
 \multicolumn{1}{c|}{} & \multicolumn{5}{c}{NDCG@10} \\
 \hline 
 Baseline & 0.346 & \textbf{0.709} & \textbf{0.348} & 0.249 & 0.612 \\
 \hline
 DeCAL 2x & \textbf{0.359} & 0.703 & \textbf{0.349} & \textbf{0.299} & 0.685 \\
 DeCAL 4x & 0.320 & 0.681 & 0.338 & 0.274 & \textbf{0.709} \\
 DeCAL 8x & 0.301 & 0.658 & 0.337 & 0.232 & 0.671 \\
 \hline
\end{tabular}%
\caption{Retrieval results for different levels of DeCAL compression and without DeCAL for 5 BEIR datasets.}
\label{table:5}
\end{center}
\end{table}

In Table~\ref{table:5}, we see that DeCAL 2x is as good as or better than the baseline, with a 4x computation savings.  DeCAL 4x performance drops off only modestly for 16x computation savings.  DeCAL 8x drops off further, but at 64x in computation savings and 8x reduction in storage versus the baseline.  

Interestingly, we see a wide dispersion of response to compression in the NDCG@10 metric.  FiQA is hurt relatively most, but COVID appears to consistently benefit even to 8x compression.  We also see that 2x compression is frequently better than no compression; we speculate that compression may be fusing subwords such that the combined information actually improves retrieval.  Doing this more intelligently may reap further benefits.  

Intuitively, compressing the query too much can hurt multi-vector retrieval since we lose conceptual granularity.  We might obtain better results at high compression using an asymmetric dual encoder with low compression on the query, but that is left for future work.

\subsection{Ablations}
In this section we focus on HotpotQA, since it showed the greatest sensitivity to compression.  We also use 2x compression since that represents the upper bound of performance under compression.  

\begin{table}[h]
\renewcommand{\arraystretch}{1.3}
\begin{center}
\begin{tabular}{c | c | c} 
 \hline
 \multicolumn{1}{c|}{} & \multicolumn{2}{c}{HotpotQA} \\
 \cline{2-3}
  & EM & F1 \\
 \hline \
 Baseline & 65.1 & 79.2 \\
 DeCAL 2x & \textbf{65.2} & \textbf{79.5} \\
 \hline
 DeCAL 2x (autoencode) & 63.9 & 78.3 \\
 DeCAL 2x (fine-tune-only) & 62.6 & 78.0 \\
 DeCAL 2x (no pooling) & 61.8 & 75.8 \\
 AttnPool 2x & 60.2 & 74.4 \\
 \hline
\end{tabular}
\caption{Ablations: SQuAD metrics for HotpotQA comparing (1) the vanilla T5 baseline, (2) DeCAL 2x,  (3) denoising replaced by autoencoding for DeCAL 2x, (4) applying DeCAL 2x during fine-tune (but not pretrain), (5) omitting token pooling for the initial latent in DeCAL 2x, and (6) AttnPool 2x.}
\label{table:6}
\end{center}
\end{table}

There are three main components of DeCAL that we can ablate:  (1) language modeling pretraining with a compressive encoder, (2) the latent-based DeCAL compressive structure itself, and (3) the token pooling when forming the initial input latent sequence.  

For (1), we can omit the denoising task entirely and first introduce compression during fine-tuning (on top of the vanilla baseline).  Alternatively, we can train compression with autoencoding (mixed 1:1 with text continuation to avoid overfitting on reconstruction) in place of denoising, as we did for CCProxy.  In Table~\ref{table:6}, we see that both of these yield a result that is notably worse than when pretrained with denoising and so can no longer hold up against the baseline.

For (2), in place of the DeCAL encoder structure, we can instead apply the commonly-used form of attention pooling by applying it on the final encoder output.  This is what we already included in Comparisons as AttnPool 2x.  Included here in context with the other ablations, we can see that this is actually the most damaging of them.  We conclude that building up the compressed representation over many layers rather than in one is critical for optimal compression.

For (3), When we use only the learnable vector \textbf{\textit{v}} and skip the initial token pooling, we see that it hurts performance significantly.  This indicates that including the representations of the input tokens plays a useful role in guiding the information transfer from the input tokens to the smaller latent sequence.

These ablations give clear support to the value of the multilayer latent-based compressive structure, denoising pretraining, and initial token pooling as important facets of the DeCAL design.

\section{Conclusion}
In this paper, we introduced DeCAL, a novel method for tokenwise compression.  We demonstrated that the compressed representations can be very usable, matching baseline at 2x compression on all tasks tested.  At a significant 8x compression, average ROUGE-L for DeCAL is only 4.1\% lower than baseline for summarization tasks, and average F1 is only 4.3\% lower for question-answering tasks.  Similarly, average NDCG@10 is only 2.9\% lower than baseline on multi-vector retrieval tasks at 8x compression, and demonstrates the versatility of DeCAL over a wide range of downstream tasks.

We extracted the key common elements of recent LLM context compressors to construct a proxy for comparison.  DeCAL outperformed on both summarization and question-answering tasks, highlighting the advantages of the combination of denoising pretraining, initial token pooling for the latent sequence, and bidirectional attention.  We also showed a strong benefit from utilizing multiple layers of computation on the compressed latent sequence alongside the full input sequence in contrast to simply attention pooling the encoder output to achieve compression.

There are many avenues for further exploration.  We found the default T5 \citep{t5} span corruption pretrain tasks worked well, but a task tailored for compression may work even better.  We have not evaluated the balance of encoder and decoder size, compressing sequences longer than 4k tokens, or more sophisticated \textbf{\textit{x}}-token to \textbf{\textit{l}}-token mappings for initial pooling to facilitate variable compression.  DeCAL may be applicable to character/byte-level models and multimodal sources.  Work remains to explore techniques to push DeCAL compression to 16x and beyond.  We have yet to try Retrieval-Augmented Generation, a common target for context compression, where we can take advantage of pre-computed compressed passages to fill large contexts.  Lastly, we can seek ways to maintain compression quality while reducing compute.

\subsubsection*{Acknowledgments}
We would like to thank the Gemini Embeddings team for assistance with multi-vector retrieval.  We also thank Joshua Ainslie for past work on sparse attention, which inspired experiments that ultimately led to the development of DeCAL.

\bibliography{panwar}

\begin{thebibliography}{30}
\providecommand{\natexlab}[1]{#1}
\providecommand{\url}[1]{\texttt{#1}}
\expandafter\ifx\csname urlstyle\endcsname\relax
  \providecommand{\doi}[1]{doi: #1}\else
  \providecommand{\doi}{doi: \begingroup \urlstyle{rm}\Url}\fi

\bibitem[Bulatov et~al.(2024)Bulatov, Kuratov, Kapushev, and Burtsev]{recurrent_memory}
Aydar Bulatov, Yuri Kuratov, Yermek Kapushev, and Mikhail~S. Burtsev.
\newblock Scaling transformer to 1m tokens and beyond with rmt, 2024.
\newblock URL \url{https://arxiv.org/abs/2304.11062}.

\bibitem[Chevalier et~al.(2023)Chevalier, Wettig, Ajith, and Chen]{chevalier2023adapting}
Alexis Chevalier, Alexander Wettig, Anirudh Ajith, and Danqi Chen.
\newblock Adapting language models to compress contexts.
\newblock In \emph{The 2023 Conference on Empirical Methods in Natural Language Processing}, 2023.
\newblock URL \url{https://openreview.net/forum?id=kp1U6wBPXq}.

\bibitem[Clark et~al.(2022)Clark, Garrette, Turc, and Wieting]{canine}
Jonathan~H. Clark, Dan Garrette, Iulia Turc, and John Wieting.
\newblock Canine: Pre-training an efficient tokenization-free encoder for language representation.
\newblock \emph{Transactions of the Association for Computational Linguistics}, 10:\penalty0 73--91, 01 2022.
\newblock ISSN 2307-387X.
\newblock \doi{10.1162/tacl_a_00448}.
\newblock URL \url{https://doi.org/10.1162/tacl\_a\_00448}.

\bibitem[Cohan et~al.(2018)Cohan, Dernoncourt, Kim, Bui, Kim, Chang, and Goharian]{arxiv_pubmed_sum}
Arman Cohan, Franck Dernoncourt, Doo~Soon Kim, Trung Bui, Seokhwan Kim, Walter Chang, and Nazli Goharian.
\newblock A discourse-aware attention model for abstractive summarization of long documents.
\newblock In Marilyn Walker, Heng Ji, and Amanda Stent (eds.), \emph{Proceedings of the 2018 Conference of the North {A}merican Chapter of the Association for Computational Linguistics: Human Language Technologies, Volume 2 (Short Papers)}, pp.\  615--621, New Orleans, Louisiana, June 2018. Association for Computational Linguistics.
\newblock \doi{10.18653/v1/N18-2097}.
\newblock URL \url{https://aclanthology.org/N18-2097/}.

\bibitem[Dai et~al.(2020)Dai, Lai, Yang, and Le]{funnel}
Zihang Dai, Guokun Lai, Yiming Yang, and Quoc~V. Le.
\newblock Funnel-transformer: filtering out sequential redundancy for efficient language processing.
\newblock In \emph{Proceedings of the 34th International Conference on Neural Information Processing Systems}, NIPS '20, Red Hook, NY, USA, 2020. Curran Associates Inc.
\newblock ISBN 9781713829546.

\bibitem[Devlin et~al.(2019)Devlin, Chang, Lee, and Toutanova]{devlin-etal-2019-bert}
Jacob Devlin, Ming-Wei Chang, Kenton Lee, and Kristina Toutanova.
\newblock {BERT}: Pre-training of deep bidirectional transformers for language understanding.
\newblock In Jill Burstein, Christy Doran, and Thamar Solorio (eds.), \emph{Proceedings of the 2019 Conference of the North {A}merican Chapter of the Association for Computational Linguistics: Human Language Technologies, Volume 1 (Long and Short Papers)}, pp.\  4171--4186, Minneapolis, Minnesota, June 2019. Association for Computational Linguistics.
\newblock \doi{10.18653/v1/N19-1423}.
\newblock URL \url{https://aclanthology.org/N19-1423/}.

\bibitem[Ge et~al.(2024)Ge, Jing, Wang, Wang, Chen, and Wei]{ge2024incontext}
Tao Ge, Hu~Jing, Lei Wang, Xun Wang, Si-Qing Chen, and Furu Wei.
\newblock In-context autoencoder for context compression in a large language model.
\newblock In \emph{The Twelfth International Conference on Learning Representations}, 2024.
\newblock URL \url{https://openreview.net/forum?id=uREj4ZuGJE}.

\bibitem[Goyal et~al.(2020)Goyal, Choudhury, Raje, Chakaravarthy, Sabharwal, and Verma]{power_bert}
Saurabh Goyal, Anamitra~Roy Choudhury, Saurabh~M. Raje, Venkatesan~T. Chakaravarthy, Yogish Sabharwal, and Ashish Verma.
\newblock Power-bert: accelerating bert inference via progressive word-vector elimination.
\newblock In \emph{Proceedings of the 37th International Conference on Machine Learning}, ICML'20. JMLR.org, 2020.

\bibitem[Guo et~al.(2022)Guo, Ainslie, Uthus, Ontanon, Ni, Sung, and Yang]{guo-etal-2022-longt5}
Mandy Guo, Joshua Ainslie, David Uthus, Santiago Ontanon, Jianmo Ni, Yun-Hsuan Sung, and Yinfei Yang.
\newblock {L}ong{T}5: {E}fficient text-to-text transformer for long sequences.
\newblock In Marine Carpuat, Marie-Catherine de~Marneffe, and Ivan~Vladimir Meza~Ruiz (eds.), \emph{Findings of the Association for Computational Linguistics: NAACL 2022}, pp.\  724--736, Seattle, United States, July 2022. Association for Computational Linguistics.
\newblock \doi{10.18653/v1/2022.findings-naacl.55}.
\newblock URL \url{https://aclanthology.org/2022.findings-naacl.55/}.

\bibitem[Jaegle et~al.(2021)Jaegle, Gimeno, Brock, Vinyals, Zisserman, and Carreira]{perceiver}
Andrew Jaegle, Felix Gimeno, Andy Brock, Oriol Vinyals, Andrew Zisserman, and Joao Carreira.
\newblock Perceiver: General perception with iterative attention.
\newblock In Marina Meila and Tong Zhang (eds.), \emph{Proceedings of the 38th International Conference on Machine Learning}, volume 139 of \emph{Proceedings of Machine Learning Research}, pp.\  4651--4664. PMLR, 18--24 Jul 2021.
\newblock URL \url{https://proceedings.mlr.press/v139/jaegle21a.html}.

\bibitem[Joshi et~al.(2017)Joshi, Choi, Weld, and Zettlemoyer]{joshi-etal-2017-triviaqa}
Mandar Joshi, Eunsol Choi, Daniel Weld, and Luke Zettlemoyer.
\newblock {T}rivia{QA}: A large scale distantly supervised challenge dataset for reading comprehension.
\newblock In Regina Barzilay and Min-Yen Kan (eds.), \emph{Proceedings of the 55th Annual Meeting of the Association for Computational Linguistics (Volume 1: Long Papers)}, pp.\  1601--1611, Vancouver, Canada, July 2017. Association for Computational Linguistics.
\newblock \doi{10.18653/v1/P17-1147}.
\newblock URL \url{https://aclanthology.org/P17-1147/}.

\bibitem[Khattab \& Zaharia(2020)Khattab and Zaharia]{colbert}
Omar Khattab and Matei Zaharia.
\newblock Colbert: Efficient and effective passage search via contextualized late interaction over bert.
\newblock In \emph{Proceedings of the 43rd International ACM SIGIR Conference on Research and Development in Information Retrieval}, SIGIR '20, pp.\  39–48, New York, NY, USA, 2020. Association for Computing Machinery.
\newblock ISBN 9781450380164.
\newblock \doi{10.1145/3397271.3401075}.
\newblock URL \url{https://doi.org/10.1145/3397271.3401075}.

\bibitem[Kudo \& Richardson(2018)Kudo and Richardson]{sentencepiece}
Taku Kudo and John Richardson.
\newblock Sentencepiece: {A} simple and language independent subword tokenizer and detokenizer for neural text processing.
\newblock \emph{CoRR}, abs/1808.06226, 2018.
\newblock URL \url{http://arxiv.org/abs/1808.06226}.

\bibitem[Li \& Liang(2021)Li and Liang]{prefix_tuning}
Xiang~Lisa Li and Percy Liang.
\newblock Prefix-tuning: Optimizing continuous prompts for generation.
\newblock \emph{CoRR}, abs/2101.00190, 2021.
\newblock URL \url{https://arxiv.org/abs/2101.00190}.

\bibitem[Nallapati et~al.(2016)Nallapati, Zhou, dos Santos, Gu{\ensuremath{\dot{}}}l{\c{c}}ehre, and Xiang]{cnnsum}
Ramesh Nallapati, Bowen Zhou, Cicero dos Santos, {\c{C}}a{\u{g}}lar Gu{\ensuremath{\dot{}}}l{\c{c}}ehre, and Bing Xiang.
\newblock Abstractive text summarization using sequence-to-sequence {RNN}s and beyond.
\newblock In Stefan Riezler and Yoav Goldberg (eds.), \emph{Proceedings of the 20th {SIGNLL} Conference on Computational Natural Language Learning}, pp.\  280--290, Berlin, Germany, August 2016. Association for Computational Linguistics.
\newblock \doi{10.18653/v1/K16-1028}.
\newblock URL \url{https://aclanthology.org/K16-1028/}.

\bibitem[Nawrot et~al.(2021)Nawrot, Tworkowski, Tyrolski, Kaiser, Wu, Szegedy, and Michalewski]{hourglass}
Piotr Nawrot, Szymon Tworkowski, Michal Tyrolski, Lukasz Kaiser, Yuhuai Wu, Christian Szegedy, and Henryk Michalewski.
\newblock Hierarchical transformers are more efficient language models.
\newblock \emph{CoRR}, abs/2110.13711, 2021.
\newblock URL \url{https://arxiv.org/abs/2110.13711}.

\bibitem[Nguyen et~al.(2016)Nguyen, Rosenberg, Song, Gao, Tiwary, Majumder, and Deng]{msmarco}
Tri Nguyen, Mir Rosenberg, Xia Song, Jianfeng Gao, Saurabh Tiwary, Rangan Majumder, and Li~Deng.
\newblock Ms marco: A human generated machine reading comprehension dataset.
\newblock November 2016.
\newblock URL \url{https://www.microsoft.com/en-us/research/publication/ms-marco-human-generated-machine-reading-comprehension-dataset/}.

\bibitem[Pagnoni et~al.(2024)Pagnoni, Pasunuru, Rodriguez, Nguyen, Muller, Li, Zhou, Yu, Weston, Zettlemoyer, Ghosh, Lewis, Holtzman, and Iyer]{pagnoni2024bytelatenttransformerpatches}
Artidoro Pagnoni, Ram Pasunuru, Pedro Rodriguez, John Nguyen, Benjamin Muller, Margaret Li, Chunting Zhou, Lili Yu, Jason Weston, Luke Zettlemoyer, Gargi Ghosh, Mike Lewis, Ari Holtzman, and Srinivasan Iyer.
\newblock Byte latent transformer: Patches scale better than tokens, 2024.
\newblock URL \url{https://arxiv.org/abs/2412.09871}.

\bibitem[Qian et~al.(2022)Qian, Lee, Duddu, Dai, Brahma, Naim, Lei, and Zhao]{aligner}
Yujie Qian, Jinhyuk Lee, Sai Meher~Karthik Duddu, Zhuyun Dai, Siddhartha Brahma, Iftekhar Naim, Tao Lei, and Vincent~Y. Zhao.
\newblock Multi-vector retrieval as sparse alignment.
\newblock \emph{CoRR}, abs/2211.01267, 2022.
\newblock \doi{10.48550/ARXIV.2211.01267}.
\newblock URL \url{https://doi.org/10.48550/arXiv.2211.01267}.

\bibitem[Qin \& Van~Durme(2023)Qin and Van~Durme]{nugget}
Guanghui Qin and Benjamin Van~Durme.
\newblock Nugget: neural agglomerative embeddings of text.
\newblock In \emph{Proceedings of the 40th International Conference on Machine Learning}, ICML'23. JMLR.org, 2023.

\bibitem[Rae et~al.(2019)Rae, Potapenko, Jayakumar, and Lillicrap]{compressive_transformer}
Jack~W. Rae, Anna Potapenko, Siddhant~M. Jayakumar, and Timothy~P. Lillicrap.
\newblock Compressive transformers for long-range sequence modelling.
\newblock \emph{CoRR}, abs/1911.05507, 2019.
\newblock URL \url{http://arxiv.org/abs/1911.05507}.

\bibitem[Raffel et~al.(2020)Raffel, Shazeer, Roberts, Lee, Narang, Matena, Zhou, Li, and Liu]{t5}
Colin Raffel, Noam Shazeer, Adam Roberts, Katherine Lee, Sharan Narang, Michael Matena, Yanqi Zhou, Wei Li, and Peter~J. Liu.
\newblock Exploring the limits of transfer learning with a unified text-to-text transformer.
\newblock \emph{J. Mach. Learn. Res.}, 21\penalty0 (1), January 2020.
\newblock ISSN 1532-4435.

\bibitem[Rau et~al.(2025)Rau, Wang, D\'{e}jean, Clinchant, and Kamps]{cocom}
David Rau, Shuai Wang, Herv\'{e} D\'{e}jean, St\'{e}phane Clinchant, and Jaap Kamps.
\newblock Context embeddings for efficient answer generation in retrieval-augmented generation.
\newblock In \emph{Proceedings of the Eighteenth ACM International Conference on Web Search and Data Mining}, WSDM '25, pp.\  493–502, New York, NY, USA, 2025. Association for Computing Machinery.
\newblock ISBN 9798400713293.
\newblock \doi{10.1145/3701551.3703527}.
\newblock URL \url{https://doi.org/10.1145/3701551.3703527}.

\bibitem[Santhanam et~al.(2022)Santhanam, Khattab, Potts, and Zaharia]{plaid}
Keshav Santhanam, Omar Khattab, Christopher Potts, and Matei Zaharia.
\newblock Plaid: An efficient engine for late interaction retrieval.
\newblock In \emph{Proceedings of the 31st ACM International Conference on Information \& Knowledge Management}, CIKM '22, pp.\  1747–1756, New York, NY, USA, 2022. Association for Computing Machinery.
\newblock ISBN 9781450392365.
\newblock \doi{10.1145/3511808.3557325}.
\newblock URL \url{https://doi.org/10.1145/3511808.3557325}.

\bibitem[{Tang} et~al.(2024){Tang}, {Wang}, {Guo}, {Tu}, {Han}, {Hu}, and {Tao}]{pruning_survey}
Yehui {Tang}, Yunhe {Wang}, Jianyuan {Guo}, Zhijun {Tu}, Kai {Han}, Hailin {Hu}, and Dacheng {Tao}.
\newblock {A Survey on Transformer Compression}.
\newblock \emph{arXiv e-prints}, art. arXiv:2402.05964, February 2024.
\newblock \doi{10.48550/arXiv.2402.05964}.

\bibitem[Tay et~al.(2022)Tay, Dehghani, Bahri, and Metzler]{transformers_survey}
Yi~Tay, Mostafa Dehghani, Dara Bahri, and Donald Metzler.
\newblock Efficient transformers: A survey.
\newblock \emph{ACM Comput. Surv.}, 55\penalty0 (6), December 2022.
\newblock ISSN 0360-0300.
\newblock \doi{10.1145/3530811}.
\newblock URL \url{https://doi.org/10.1145/3530811}.

\bibitem[Thakur et~al.(2021)Thakur, Reimers, R{\"u}ckl{\'e}, Srivastava, and Gurevych]{thakur2021beir}
Nandan Thakur, Nils Reimers, Andreas R{\"u}ckl{\'e}, Abhishek Srivastava, and Iryna Gurevych.
\newblock {BEIR}: A heterogeneous benchmark for zero-shot evaluation of information retrieval models.
\newblock In \emph{Thirty-fifth Conference on Neural Information Processing Systems Datasets and Benchmarks Track (Round 2)}, 2021.
\newblock URL \url{https://openreview.net/forum?id=wCu6T5xFjeJ}.

\bibitem[Vaswani et~al.(2017)Vaswani, Shazeer, Parmar, Uszkoreit, Jones, Gomez, Kaiser, and Polosukhin]{Vaswani_NIPS2017_3f5ee243}
Ashish Vaswani, Noam Shazeer, Niki Parmar, Jakob Uszkoreit, Llion Jones, Aidan~N Gomez, \L~ukasz Kaiser, and Illia Polosukhin.
\newblock Attention is all you need.
\newblock In I.~Guyon, U.~Von Luxburg, S.~Bengio, H.~Wallach, R.~Fergus, S.~Vishwanathan, and R.~Garnett (eds.), \emph{Advances in Neural Information Processing Systems}, volume~30. Curran Associates, Inc., 2017.
\newblock URL \url{https://proceedings.neurips.cc/paper_files/paper/2017/file/3f5ee243547dee91fbd053c1c4a845aa-Paper.pdf}.

\bibitem[Yang et~al.(2018)Yang, Qi, Zhang, Bengio, Cohen, Salakhutdinov, and Manning]{yang-etal-2018-hotpotqa}
Zhilin Yang, Peng Qi, Saizheng Zhang, Yoshua Bengio, William Cohen, Ruslan Salakhutdinov, and Christopher~D. Manning.
\newblock {H}otpot{QA}: A dataset for diverse, explainable multi-hop question answering.
\newblock In Ellen Riloff, David Chiang, Julia Hockenmaier, and Jun{'}ichi Tsujii (eds.), \emph{Proceedings of the 2018 Conference on Empirical Methods in Natural Language Processing}, pp.\  2369--2380, Brussels, Belgium, October-November 2018. Association for Computational Linguistics.
\newblock \doi{10.18653/v1/D18-1259}.
\newblock URL \url{https://aclanthology.org/D18-1259/}.

\bibitem[Zhu et~al.(2021)Zhu, Liu, Mei, and Zeng]{zhu-etal-2021-mediasum}
Chenguang Zhu, Yang Liu, Jie Mei, and Michael Zeng.
\newblock {M}edia{S}um: A large-scale media interview dataset for dialogue summarization.
\newblock In Kristina Toutanova, Anna Rumshisky, Luke Zettlemoyer, Dilek Hakkani-Tur, Iz~Beltagy, Steven Bethard, Ryan Cotterell, Tanmoy Chakraborty, and Yichao Zhou (eds.), \emph{Proceedings of the 2021 Conference of the North American Chapter of the Association for Computational Linguistics: Human Language Technologies}, pp.\  5927--5934, Online, June 2021. Association for Computational Linguistics.
\newblock \doi{10.18653/v1/2021.naacl-main.474}.
\newblock URL \url{https://aclanthology.org/2021.naacl-main.474/}.

\end{thebibliography}
\bibliographystyle{iclr2026_conference}

\end{document}